%% file: cleverhans.tex
\documentclass[]{article}

\usepackage{hyperref}
\usepackage{authblk}
\usepackage{amsmath,amssymb,amsfonts,amsthm}

\title{Technical Report on the \texttt{cleverhans} v2.1.0 Adversarial Examples Library}

\author[1,3]{Nicolas Papernot\thanks{ngp5056@cse.psu.edu}}
\author[5,3]{Fartash Faghri}
\author[2,3]{Nicholas Carlini}
\author[3]{Ian Goodfellow\thanks{goodfellow@google.com}}
\author[4]{Reuben Feinman}
\author[3]{Alexey Kurakin}
\author[6]{Cihang Xie}
\author[7]{Yash Sharma}
\author[3]{Tom Brown}
\author[3]{Aurko Roy}
\author[8]{Alexander Matyasko}
\author[9]{Vahid Behzadan}
\author[10]{Karen Hambardzumyan}
\author[6]{Zhishuai Zhang}
\author[11]{Yi-Lin Juang}
\author[5]{Zhi Li}
\author[1]{Ryan Sheatsley}
\author[12]{Abhibhav Garg}
\author[13]{Jonathan Uesato}
\author[14]{Willi Gierke}
\author[15]{Yinpeng Dong}
\author[3]{David Berthelot}
\author{Paul Hendricks}
\author[16]{Jonas Rauber}
\author[17]{Rujun Long}
\author[1]{Patrick McDaniel\thanks{mcdaniel@cse.psu.edu}}
\affil[1]{Pennsylvania State University}
\affil[2]{UC Berkeley}
\affil[3]{Google Brain}
\affil[4]{Symantec}
\affil[5]{University of Toronto}
\affil[6]{Johns Hopkins}
\affil[7]{The Cooper Union}
\affil[8]{Nanyang Technological University}
\affil[9]{Kansas State}
\affil[10]{YerevaNN}
\affil[11]{NTUEE}
\affil[12]{IIT Delhi}
\affil[13]{MIT}
\affil[14]{Hasso Plattner Institute}
\affil[15]{National Tsing Hua University}
\affil[16]{IMPRS}
\affil[17]{0101.AI}

\date{}

\begin{document}

\maketitle
\newpage
\input{abstract}

\input{introduction}

\input{core}

\input{benchmark}
\input{version}

\section{Acknowledgments}
The format of this report was in part inspired
by~\cite{2016arXiv160502688short}. Nicolas Papernot is supported by a Google PhD
Fellowship in Security. Research was sponsored by the Army Research Laboratory
and was accomplished under Cooperative Agreement Number W911NF-13-2-0045 (ARL
Cyber Security CRA). The views and conclusions contained in this document are
those of the authors and should not be interpreted as representing the official
policies, either expressed or implied, of the Army Research Laboratory or the
U.S. Government. The U.S. Government is authorized to reproduce and distribute
reprints for Government purposes notwithstanding any copyright notation here on.

\bibliography{cleverhans.bbl}
\bibliographystyle{plain}

\end{document}

%% file: abstract.tex
\begin{abstract}
\texttt{cleverhans} is a software library that provides standardized reference
implementations of \emph{adversarial example} construction techniques and
\emph{adversarial training}. The library may be used to develop more robust
machine learning models and to provide standardized benchmarks of models'
performance in the adversarial setting. Benchmarks constructed without a
standardized implementation of adversarial example construction are not
comparable to each other, because a good result may indicate a robust model or
it may merely indicate a weak implementation of the adversarial example
construction procedure.

This technical report is structured as follows. Section~\ref{sec:introduction}
provides an overview of adversarial examples in machine learning and of the
\texttt{cleverhans} software. Section~\ref{sec:core} presents the core
functionalities of the library: namely the attacks based on adversarial examples
and defenses to improve the robustness of machine learning models to these
attacks. Section~\ref{sec:benchmark} describes how to report benchmark results
using the library. Section~\ref{sec:version} describes the versioning system.

\end{abstract}

%% file: introduction.tex
\section{Introduction}
\label{sec:introduction}

Adversarial examples are inputs crafted by making slight perturbations to
legitimate inputs with the intent of misleading machine learning
models~\cite{szegedy2013intriguing}. The perturbations are designed to be small
in magnitude, such that a human observer would not have difficulty processing
the resulting input. In many cases, the perturbation required to deceive a
machine learning model is so small that a human being may not be able to
perceive that anything has changed, or even so small that an 8-bit
representation of the input values does not capture the perturbation used to
fool a model that accepts 32-bit inputs. We invite readers unfamiliar with the
concept to the detailed presentation
in~\cite{szegedy2013intriguing,goodfellow2014explaining,papernot2016limitations,biggio2013evasion}. Although completely effective defenses have yet to be proposed, the most successful to date is adversarial training~\cite{szegedy2013intriguing,goodfellow2014explaining}.
Different sources of adversarial examples used in the training process
can make adversarial training more effective; as of this writing, to the
best of our knowledge, the most effective version of adversarial training
on ImageNet is ensemble adversarial training~\cite{tramer2017ensemble}
and the most effective version on MNIST is the basic iterative method~\cite{kurakin2016adversarial}
applied to randomly chosen starting points~\cite{madry2017towards}.

The \texttt{cleverhans} library provides reference implementations of the
attacks, which are intended for use for two purposes. First, machine learning
developers may construct robust models by using adversarial training, which
requires the construction of adversarial examples during the training procedure.
Second, we encourage researchers who report the accuracy of their models in the
adversarial setting to use the standardized reference implementation provided by
\texttt{cleverhans}. Without a standard reference implementation, different
benchmarks are not comparable---a benchmark reporting high accuracy might
indicate a more robust model, but it might also indicate the use of a weaker
attack implementation. By using \texttt{cleverhans}, researchers can be assured
that a high accuracy on a benchmark corresponds to a robust model.

Implemented in
TensorFlow~\cite{abadi2016tensorflow},
\texttt{cleverhans} is designed as a tool
to help developers add defenses against adversarial examples
to their models
and benchmark the robustness of their models to adversarial
examples.
The interface for \texttt{cleverhans} is designed to accept
models implemented using any model framework
( such as
Keras~\cite{chollet2015keras})
or implemented without any specific model abstraction.

% IG thinks we probably do not need to say this, but is less sure about that
% than some of his other edits.
%
% As \texttt{cleverhans} is implemented using existing numerical computation
% libraries, it benefits from the computational speed-ups offered by
% the use of GPUs without requiring changes to
% the code, so long as the user has correctly configured the underlying
% numerical computation library.

The \texttt{cleverhans} library is a collaboration
is free, open-source software, licensed under
the MIT license. The project is available online through
GitHub\footnote{\url{https://github.com/openaicleverhans}}. The main
communication channel for developers of the library is a mailing list, whose
discussions are publicly available
online\footnote{\url{https://groups.google.com/group/cleverhans-dev}}.

%% file: core.tex
\section{Core functionalities}
\label{sec:core}

The library's package is organized by modules. The most important modules are:
\begin{itemize} \item \texttt{attacks}: contains the \texttt{Attack}
      class, defining the interface used by all CleverHans attacks,
      as well as implementations of several specific attacks.

\item \texttt{model}: contains the \texttt{Model} class, which is
      a very lightweight class defining a simple interface that
      models should implement in order to be compatible with
      \texttt{Attack}. CleverHans includes a \texttt{Model} implementation
      for Keras \texttt{Sequential} models and examples of \texttt{Model}
      implementations for TensorFlow models that are not implemented
      using any modeling framework library.

\end{itemize}

In the following, we describe some of the research results behind the
implementations made in \texttt{cleverhans}.

\subsection{Attacks}

Adversarial example crafting algorithms implemented in \texttt{cleverhans} take
a model, and an input, and return the corresponding adversarial example. Here
are the algorithms currently implemented in the \texttt{attacks} module.

\subsubsection{L-BFGS Method}

The L-BFGS method was introduced by Szegedy et al.~\cite{szegedy2013intriguing}. It aims to solve the following box-constrained optimization problem: 
\begin{align}
\label{eqn_EAD_formulation_1}
 &\textnormal{minimize}~~\|x_0 - x\|_2^2 \nonumber \\
 &\textnormal{such that}~~C(x) = l \nonumber \\
 &\textnormal{where}~~x \in [0,1]^p 
\end{align}
The computation is approximated by using box-constrained L-BFGS optimization. 

\subsubsection{Fast Gradient Sign Method} \label{sec:fgsm}

The fast gradient sign method (FGSM) was introduced by Goodfellow et
al.~\cite{goodfellow2014explaining}. The intuition behind the attack is to
linearize the cost function $J$ used to train a model $f$ around the
neighborhood of the training point $\vec{x}$ that the adversary wants to force
the misclassification of. The resulting adversarial example $\vec{x}^*$
corresponding to input $\vec{x}$ is computed as follows: \begin{equation}
\vec{x}^* \leftarrow x + \varepsilon \cdot \nabla_{\vec{x}}J(f, \theta, \vec{x})
\end{equation} where $\varepsilon$ is a parameter controlling the magnitude of
the perturbation introduced. Larger values increase the likelihood that
$\vec{x}^*$ will be misclassified by $f$, but make the perturbation easier to
detect by a human.

The fast gradient sign method is available by calling \texttt{attacks.fgsm()}
The implementation defines the necessary graph elements and returns a tensor,
which once evaluated holds the value of the adversarial example corresponding to
the input provided. The implementation is parameterized by the parameter
$\varepsilon$ introduced above. It is possible to configure the method to clip
adversarial examples so that they are constrained to be part of the expected
input domain range.

\subsubsection{Carlini-Wagner Attack}

The Carlini-Wagner (C\&W) attack was introduced by Carlini et al.~\cite{carlini2016towards}. Inspired by~\cite{szegedy2013intriguing}, the authors formulate finding adversarial examples as an optimization problem; find some small change $\delta$ that can be made to an input $x$ that will change its classification, but so that the result is still in the valid range. They instantiate the distance metric with an $L_{p}$ norm, define a success function $f$ such that $f(x + \delta)\leq0$ if and only if the model misclassifies, and minimize the sum with a trade-off constant `c'. `c' is chosen by modified binary search, the box constraint is resolved by applying a change-of-variables, and the Adam~\cite{adam} optimizer is used to solve the optimization instance. 

The attack has been shown to be quite powerful~\cite{carlini2016towards, carlini2017detection}, however this power comes at the cost of speed, as this attack is often much slower than others. The attack can be sped up by fixing `c' (instead of performing modified binary search). 

The Carlini-Wagner attack is available by instantiating the attack object with \texttt{attacks.CarliniWagnerL2} and then calling the \texttt{generate()} function. This generates the symbolic graph and returns a tensor, which once evaluated holds the value of the adversarial example corresponding to the input provided. As the name suggests, the $L_{p}$ norm used in the implementation is $L_2$. The attack is controlled by a number of parameters, namely the confidence, which defines the margin between logit values necessary to succeed, the learning rate (step-size), the number of binary search steps, the number of iterations per binary search step, and the initial `c' value.   

\subsubsection{Elastic Net Method}

The Elastic Net Method (EAD) was introduced by Chen et al.~\cite{chen2017ead}. Inspired by the C\&W attack~\cite{carlini2016towards}, finding adversarial examples is formulated as an optimization problem. The same loss function as used by the C\&W attack is adopted, however instead of performing $L_2$ regularization, elastic-net regularization is performed, with $\beta$ controlling the trade-off between $L_1$ and $L_2$. The iterative shrinkage-thresholding algorithm (ISTA)~\cite{beck2009fast}. ISTA can be viewed as a regular first-order optimization algorithm with an additional shrinkage-thresholding step on each iteration. 

Notably, the C\&W $L_2$ attack becomes a special case of the EAD formulation, with $\beta = 0$. However, one can view EAD as a robust version of the C\&W method, as the ISTA operation shrinks a value of the adversarial example if the deviation to the original input is greater than $\beta$, and leaves the value unchanged if the deviation is less than $\beta$. Empirical results support this claim, demonstrating the attack's ability to bypass strong detection schemes and succeed against robust adversarially trained models while still producing adversarial examples with minimal visual distortion~\cite{chen2017ead,sharma2017madry,sharma2018feature,lu2018magnet}. 

The Elastic Net Method is available by instantiating the attack object with \texttt{attacks.ElasticNetMethod} and then calling the \texttt{generate()} function. This generates the symbolic graph and returns a tensor, which once evaluated holds the value of the adversarial example corresponding to the input provided. The attack is controlled by a number of parameters, most of which are shared with the C\&W attack, namely the confidence, which defines the margin between logit values necessary to succeed, the learning rate (step-size), the number of binary search steps, the number of iterations per binary search step, and the initial `c' value. Additional parameters include $\beta$, the elastic-net regularization constant, and the decision rule, whether to choose successful adversarial examples with minimal $L_1$ or elastic-net distortion.   

\subsubsection{Basic Iterative Method} \label{sec:bim}

The basic iterative method (BIM) was introduced by Kurakin et al.~\cite{kurakin2016adversarial}, and extends the ``fast'' gradient method by applying it multiple times with small step size, clipping values of intermediate results after each step to ensure that they are in an $\varepsilon$-neighborhood of the original input. 

The basic iterative method is available by instantiating the attack object with \texttt{attacks.BasicIterativeMethod} and then calling the \texttt{generate()} function. This generates the symbolic graph and returns a tensor, which once evaluated holds the value of the adversarial example corresponding to the input provided. The attack is parameterized by $\varepsilon$, alike the fast gradient method, but also by the step-size for each attack iteration and the number of attack iterations.

\subsubsection{Projected Gradient Descent}

The projected gradient descent (PGD) attack was introduced by Madry et al.~\cite{madry2017towards}. The authors state that the basic iterative method (BIM)~\cite{kurakin2016adversarial} is essentially projected gradient descent on the negative loss function. To explore the loss landscape further, PGD is re-started from many points in the $L_\infty$ balls around the input examples. 

PGD is available by instantiating the attack object with \texttt{attacks.MadryEtAl} and then calling the \texttt{generate()} function. This generates the symbolic graph and returns a tensor, which once evaluated holds the value of the adversarial example corresponding to the input provided. PGD shares many parameters with BIM, such as $\varepsilon$, the step-size for each attack iteration, and the number of attack iterations. An additional parameter is a boolean which specifies whether or not to add an initial random perturbation. 

\subsubsection{Momentum Iterative Method}

The momentum iterative method (MIM) was introduced by Dong et al.~\cite{dong2017momentum}. It is a technique for accelerating
gradient descent algorithms by accumulating a velocity
vector in the gradient direction of the loss function across
iterations. BIM with incorporated momentum applied to an ensemble of models won first place in both the NIPS 2017 Non-Targeted and Targeted Adversarial Attack Competitions~\cite{nipscompetition}.

The momentum iterative method is available by instantiating the attack object with \texttt{attacks.MomentumIterativeMethod} and then calling the \texttt{generate()} function. This generates the symbolic graph and returns a tensor, which once evaluated holds the value of the adversarial example corresponding to the input provided. MIM shares many parameters with BIM, such as $\varepsilon$, the step-size for each attack iteration, and the number of attack iterations. An additional parameter is a decay factor which can be applied to the momentum term.

\subsubsection{Jacobian-based Saliency Map Approach}

The Jacobian-based saliency map approach (JSMA) was introduced by 
Papernot et al.~\cite{papernot2016limitations}. The method 
iteratively perturbs features of the input that have large adversarial saliency
scores. Intuitively, this score reflects the adversarial goal of 
taking a sample away from its source class towards a chosen 
target class. 

First, the adversary 
computes the Jacobian of the model and evaluates it in the 
current input: this returns a matrix $\left[\frac{\partial f_j}{\partial x_i}(\vec{x})\right]_{i,j}$ where component $(i,j)$ is the derivative of class $j$ with respect to input feature $i$.
To compute the adversarial saliency map, the adversary then 
computes the following for each input feature $i$:
\begin{equation}
 S(\vec{x},t)[i] = \left\lbrace
 \begin{array}{c}
 0  \mbox{ if }   \frac{\partial f_{t}(\vec{x})}{\partial \vec{x}_i}<0  \mbox{ or } \sum_{j\neq t} \frac{\partial f_{j}(\vec{x})}{\partial \vec{x}_i}>0\\
 \left(  \frac{\partial f_{t}(\vec{x})}{\partial \vec{x}_i}\right)  \left| \sum_{j\neq t} \frac{\partial f_{j}(\vec{x})}{\partial \vec{x}_i}\right| \mbox{ otherwise}
 \end{array}\right.
\end{equation}
where $t$ is the target class that the adversary wants the machine learning model to assign. The adversary then selects the 
input feature $i$ with the largest saliency score $S(\vec{x},t)[i]$ and increases its value\footnote{In the original paper and the \texttt{cleverhans} implementation, input 
	features are selected by pairs using the same heuristic.}. 
The process is repeated until misclassification in the target class is achieved
or the maximum number of perturbed features has been reached. 

In \texttt{cleverhans}, the Jacobian-based saliency map approach may be called with \texttt{attacks.jsma()}. The implementation 
returns the adversarial example directly, as well as whether 
the target class was achieved or not, and how many input
features were perturbed.

\subsubsection{DeepFool}

DeepFool was introduced by Moosavi-Dezfooli et al.~\cite{moosavi2015deepfool}. Unlike most of the attacks described here, it cannot be used in the targeted case, where the attacker specifies what target class the model should classify the adversarial example as. It can only be used in the non-targeted case, where the attacker can only ensure that the the model classifies the adversarial example in a class different from the original.

Inspired by the fact that the corresponding separating hyperplanes in linear classifiers indicate the decision boundaries of each class, DeepFool aims to find the least distortion (in terms of euclidean distance) leading to misclassification by projecting the input example to the closest separating hyperplane. An approximate iterative algorithm is proposed for attacking neural networks in order to tackle its inherent nonlinearities. 

DeepFool is available by instantiating the attack object with \texttt{attacks.DeepFool} and then calling the \texttt{generate()} function. This generates the symbolic graph and returns a tensor, which once evaluated holds the value of the adversarial example corresponding to the input provided. DeepFool has a few parameters, such as the number of classes to test against, a termination criterion to prevent vanishing updates, and the maximum number of iterations. 

\subsubsection{Feature Adversaries}

Feature Adversaries were introduced by Sabour et al.~\cite{sabour2015feature}. Instead of solely considering adversaries which disrupt classification, termed \textit{label adversaries}, the authors considered adversarial examples which are confused with other examples not just in class label, but in their internal representations as well. Such examples are generated by \textit{feature adversaries}. 

Such feature adversarial examples are generated by minimizing the euclidean distance between the internal deep representation (at a specified layer) while constraining the distance between the input and adversarial example in terms of $L_\infty$ to be less than $\delta$. The optimization is conducted using box-constrained L-BFGS. 

Feature adversaries are available by instantiating the attack object with \texttt{attacks.FastFeatureAdversaries} and then calling the \texttt{generate()} function. This generates the symbolic graph and returns a tensor, which once evaluated holds the value of the adversarial example corresponding to the input provided. The implementation is parameterized by the following set of parameters: $\varepsilon$, the step-size for each attack iteration, the number of attack iterations, and the layer to target.

\subsubsection{SPSA}

Simultaneous perturbation stochastic approximation (SPSA) was introduced by Uesato et al.~\cite{uesato2018spsa}. SPSA is a gradient-free optimization method, which is useful when the model is non-differentiable, or more generally, the gradients do not point in useful directions. Gradients are approximated using finite difference estimates~\cite{chen2017zoo} in random directions. 

SPSA is available by instantiating the attack object with \texttt{attacks.SPSA} and then calling the \texttt{generate()} function. This generates the symbolic graph and returns a tensor, which once evaluated holds the value of the adversarial example corresponding to the input provided. The implementation is parameterized by the following set of parameters: $\varepsilon$, the number of optimization steps, the learning rate (step-size), and the perturbation size used for the finite difference approximation.

\subsection{Defenses}

The intuition behind defenses against adversarial examples is to make the model
smoother by limiting its sensitivity to small perturbations of its inputs (and
therefore making adversarial examples harder to craft). Since all defenses
currently proposed modify the learning algorithm used to train the model, we
implement them in the modules of \texttt{cleverhans} that contain the functions
used to train models. In module \texttt{utils\_tf}, the following defenses are
implemented.

\subsubsection{Adversarial training}

The intuition behind adversarial
training~\cite{szegedy2013intriguing,goodfellow2014explaining} is to inject
adversarial examples during training to improve the generalization of the
machine learning model. To achieve this effect, the training function
\texttt{tf\_model\_train()} implemented in module \texttt{utils\_tf} can be
given the tensor definition for an adversarial example: e.g., the one returned
by the method described in Section~\ref{sec:fgsm}. When such a tensor is given,
the training algorithm modifies the loss function used to optimize the model
parameters: it is in that case defined as the average between the loss for
predictions on legitimate inputs and the loss for predictions made on
adversarial examples. The remainder of the training algorithm is left unchanged.

%% file: benchmark.tex
\section{Reporting Benchmark Results}
\label{sec:benchmark}

This section provides instructions for how to preprare and report benchmark
results.

When comparing against previously published benchmarks, it is best to to use the
same version of \texttt{cleverhans} as was used to produce the previous
benchmarks. This minimizes the possibility that an undetected change in behavior
between versions could cause a difference in the output of the benchmark
results.

When reporting new results that are not directly compared to previous work, it
is best to use the most recent versioned release of \texttt{cleverhans}.

In all cases, it is important to report the version number of
\texttt{cleverhans}.

In addition to this information, one should also report which attack methods
were used, and the values of any configuration parameters used for these
attacks.

For example, you might report ``We benchmarked the robustness of our method to
adversarial attack using v2.1.0 of CleverHans (Papernot et al. 2018).
On a test set modified by \texttt{fgsm} with \texttt{eps} of 0.3, we obtained a
test set accuracy of 97.9\%.''

The library does not provide specific test datasets or data preprocessing. End
users are responsible for appropriately preparing the data in their specific
application areas, and for reporting sufficient information about the data
preprocessing and model family to make benchmarks appropriately comparable.

%% file: version.tex
\section{Versioning}
\label{sec:version}

Because one of the goals of \texttt{cleverhans} is to provide a basis for
reproducible benchmarks, it is important that the version numbers provide useful
information. The library uses semantic
versioning,\footnote{\url{http://semver.org/}} meaning that version numbers take
the form of MAJOR.MINOR.PATCH.

The PATCH number increments whenever backwards-compatible bug fixes are made.
For the purpose of this library, a bug is not considered backwards-compatible if
it changes the results of a benchmark test. The MINOR number increments whenever
new features are added in a backwards-compatible manner. The MAJOR number
increments whenever an interface changes.

Any time a bug in CleverHans affects the accuracy of any performance number
reported as a benchmark result, we consider fixing the bug to constitute an
API change (to the interface mapping from the specification of a benchmark
experiment to the reported performance) and increment the MAJOR version number
when we make the next release.
For this reason, when writing academic articles, it is important to compare
CleverHans benchmark results that were produced with the same MAJOR version
number.
Release notes accompanying each revision
indicate whether an increment to the MAJOR number invalidates earlier
benchmark results or not.

Release notes for each version are available at
\url{https://github.com/tensorflow/cleverhans/releases}